\pdfoutput=1

\documentclass[11pt]{article}

\usepackage{acl}

\usepackage{times}
\usepackage{latexsym}

\usepackage[T1]{fontenc}

\usepackage{bm}
\usepackage{bbm}
\usepackage{amssymb} 

\usepackage[utf8]{inputenc}

\usepackage{microtype}

\usepackage{amsmath}
\usepackage{graphicx}
\usepackage{float}
\usepackage{subcaption}
\usepackage{amsfonts}
\usepackage{multicol}
\usepackage[bottom]{footmisc}
\usepackage{booktabs}
\usepackage{multirow}

\usepackage{pifont}

\newcommand{\argmax}[1]{\underset{#1}{\text{argmax}} \;}

\title{Analysis of the Cambridge Multiple-Choice Questions Reading Dataset with a Focus on Candidate Response Distribution}

\author{Adian Liusie$^{1}$, Vatsal Raina$^{1}$, Andrew Mullooly$^{2}$, Kate Knill$^{1}$, Mark J. F. Gales$^{1}$ \\
  $^{1}$ALTA Institute, Engineering Department, University of Cambridge, UK \\
  $^{2}$Cambridge University Press \& Assessment, Cambridge, UK \\ 
  \texttt{\{al826,vr311,mjfg\}@cam.ac.uk} \\
  }

\begin{document}
\pagenumbering{arabic}
\pagestyle{plain}
\maketitle
\begin{abstract}
Multiple choice exams are widely used to assess candidates across a diverse range of domains and tasks. To moderate question quality, newly proposed questions often pass through pre-test evaluation stages before being deployed into real-world exams. Currently, this evaluation process is manually intensive, which can lead to time lags in the question development cycle. Streamlining this process via automation can significantly enhance efficiency, however, there's a current lack of datasets with adequate pre-test analysis information. In this paper we analyse a subset of the public Cambridge Multiple-Choice Questions Reading Database released by Cambridge University Press \& Assessment; a multiple-choice comprehension dataset of questions at different target levels, with corresponding candidate selection distributions. We introduce the task of candidate distribution matching, propose several evaluation metrics for the task, and demonstrate that automatic systems trained on RACE++ can be leveraged as baselines for our task. We further demonstrate that these automatic systems can be used for practical pre-test evaluation tasks such as detecting underperforming distractors, where our detection systems can automatically identify poor distractors that few candidates select. 
\end{abstract}

\section{Introduction}
Multiple choice tests are used globally to assess candidates, and have been employed in diverse settings such as university admissions, job screening and qualifications accreditation \citep{AldersonJ.Charles.2000AR}. With their crucial influence on real-world decisions, it is vital that multiple-choice exams reliably assess candidate's ability. To ensure this, test makers often employ a detailed evaluation process where new questions are first internally reviewed and evaluated on pre-test candidates. This way test-makers can approve suitable questions to be deployed into real exams, while flagging any questions that should be further edited and improved. The question development cycle is, though, currently very manual and effort-intensive, which introduces time lags and costs. It would be highly beneficial to automate this process with NLP systems, however, most existing multiple choice datasets \citep{pmlr-v101-liang19a, Yu2020ReClorAR, Huang2019CosmosQM, sun2019dream} only focus on assessing system ability, where only ground-truth are provided. Few existing datasets provide candidate response distributions over questions, limiting the scope of possible solutions for automating pre-test evaluation.

\begin{figure}[t]
    \centering
    \includegraphics[width=\columnwidth]{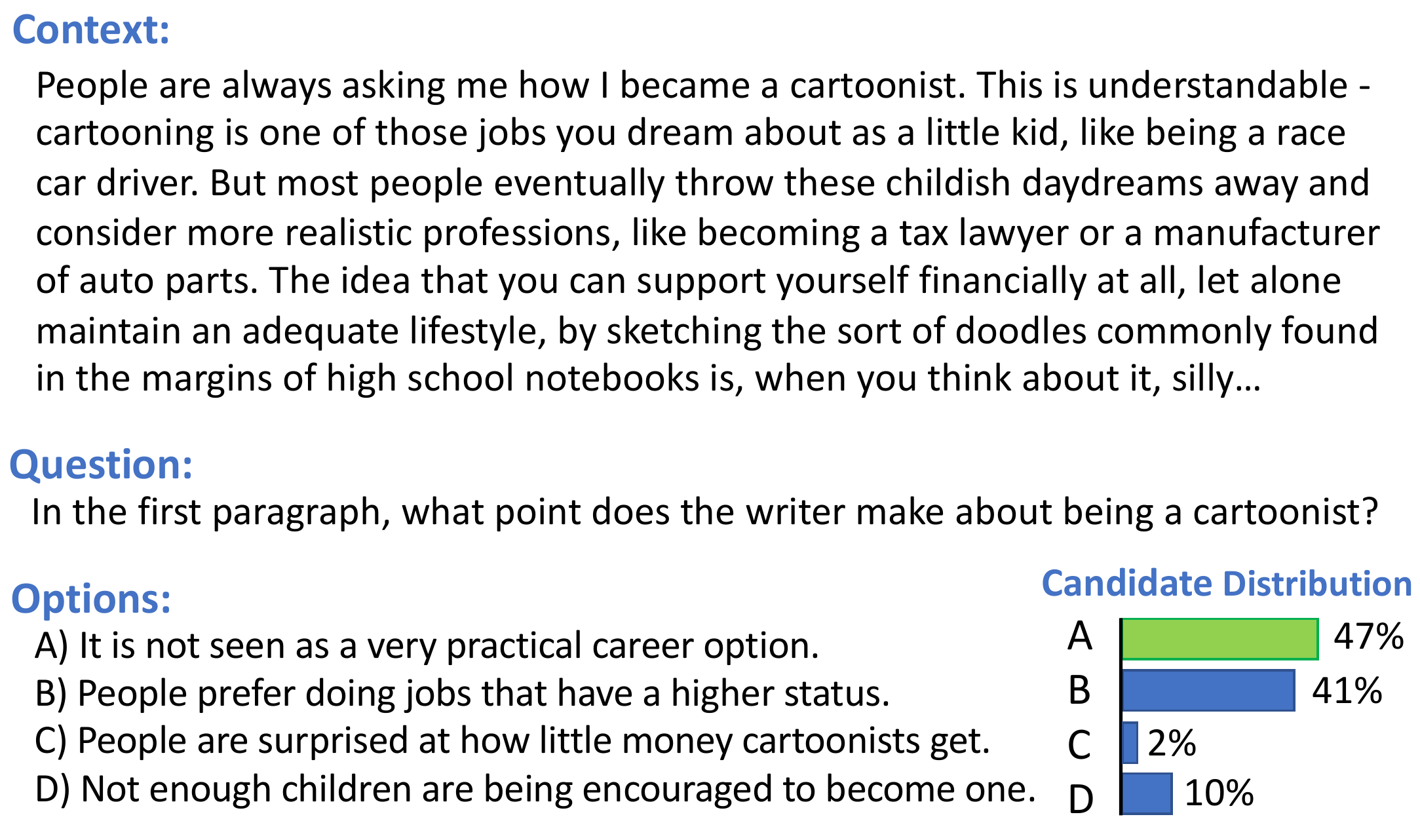}
    \caption{Sample from the CMCQRD. Each example has context (which has been truncated in the above example), question and options, and correct answer labels and distribution of candidate's selections.}
    \label{fig:example_question}
\end{figure}

In this paper, we analyse a subset of the Cambridge Multiple Choice Questions Reading Dataset (CMCQRD) \citep{CMCQRD-2023}, a  public multiple-choice reading comprehension dataset recently released by Cambridge University Press \& Assessment. CMCQRD consists of 772 items (questions) over 120 tasks targeted at various proficiency levels. Here we focus on questions for which meta-information of real candidate distribution of selected options is available, and use this 448 question subset of questions from the CMCQRD database as an evaluation set. This dataset enables the proposed task of candidate distribution matching, where given a new question, the objective is for an automatic system to predict the overall candidate selection distribution. In this paper, we propose evaluation metrics to measure the distance between candidate distributions and introduce various baseline approaches that leverage existing automatic multiple-choice answering systems.

\section{Related Data}
Numerous public multiple-choice datasets enable assessment for a range of attributes, such as scientific knowledge \citep{lu2022learn}, commonsense reasoning \citep{talmor2018commonsenseqa}, story continuation \citep{mostafazadeh2017lsdsem} and logical reasoning \citep{liu2020logiqa}. For multiple-choice reading comprehension, large-scale datasets  such as RACE++ \citep{pmlr-v101-liang19a}, COSMOSQA \citep{Huang2019CosmosQM} and ReClor \citep{Yu2020ReClorAR} are available. For the above datasets, only ground truth labels are available; though this enables one to assess system performance, no dataset currently enables automating pre-testing and evaluating prediction of candidate distribution. 

\section{Cambridge Multiple-Choice Questions Reading Dataset}
In this paper, we will only consider the subset of CMCQRD~\cite{CMCQRD-2023} with associated candidate distributions. This subset (which for convenience we will continue describing as CMCQRD) consists of 448 questions, from 78 unique contexts, taken from multiple-choice reading comprehension exams aimed at non-native speakers wishing to gain English accreditation. The questions range in difficulty between B1-C2 on the Common European Framework of Reference for Languages (CEFR), where each exam assesses comprehension ability on the \textit{context}. Questions statistics are shown in Table \ref{tab:camchoice_split}. 

\begin{table}[H]
    \centering
    \small
    \begin{tabular}{c|ccccc}
        \toprule
                    & \multicolumn{5}{c}{level} \\
                    & B1  & B2  & C1 & C2 & All \\
        \midrule
        Contexts    & 23  & 37  & 12 & 6  & 78  \\
        Questions   & 115 & 222 & 72 & 39 & 448 \\
        \bottomrule
    \end{tabular}
    \caption{Data splits for CMCQRD data (for the examples with associated candidate distributions).}
    \label{tab:camchoice_split}
\end{table}

\noindent Pretesting plays a crucial role in quality assurance; by trialling questions on L1s (native English speakers) as well as a select group of students chosen for their proficiency level, one can gain statistical information on the performance of each question. If either the quantitative statistical data or qualitative expert judgement indicates an issue, the question is not used in live tests and subsequent re-pretesting is done. Classical test theory is used to analyse item performance \citep{corrigan2015, elliott2015}, where the option-level facility and discrimination values help identify not only whether a particular question is performing as expected, but also what may cause under-performance in the question. The CMCQRD provides the candidate distribution of selected options where each question has been answered by at least 100 candidates, as well as the option-level discrimination values. 

\section{Candidate Distribution Matching Task}
In this section we introduce machine reading comprehension (MRC) systems, the task of candidate distribution matching (where MRC systems can be leveraged as initial baselines), and how candidate distribution matching can be evaluated. 

\label{sec:mcrc}
\subsection{Multiple-choice comprehension}

Machine reading comprehension (MRC) systems aim to automatically perform reading comprehension. In this work, we use the standard multiple-choice answering architecture from \citet{Yu2020ReClorAR, raina-gales-2022-answer}, where the MRC system returns a probability for each option. In standard settings probabilities are interpreted as model confidence. For the downstream application of candidate distribution matching, however, these probabilities can instead be used to represent the fraction of selected options over a population. In this use-case, we consider two useful system metrics: the mode accuracy and true class probability. 
\vspace{2mm}

\noindent \textbf{Mode Accuracy} MRC systems model the probability of the option given the input $P_{\theta}(y|x)$, where the input $x=(C, Q, \{O\})$ is the combination of context, question and options. In deployment, the predicted choice is assumed to be the option with the highest associated probability. Mode accuracy is equivalent to the traditional definition of accuracy, though when used to model a distribution of candidates, the mode accuracy represents how often the most selected option is the correct answer.
\begin{align}
    \hat{y} &= \argmax{y} P_{\theta}(y|x) \\
    \text{accuracy} &= \mathbb{E}_{x} \{ \delta(\hat{y},\; y_{ans}) \} \label{eq:accuracy}
\end{align} 
\noindent\textbf{True Class Probability} The true class probability instead looks at how much probability mass is on the correct answer, which represents the fraction of candidates that get questions correct.
\begin{equation}
    \text{true class prob.} =\mathbb{E}_{x} \{ P(y_{ans}|x) \} \label{eq:true_class}
\end{equation}

\subsection{Candidate distribution matching}
Let $P_c(y|x, \alpha)$ represent the true candidate answer selection distribution, with $\alpha \in \mathbb{R}$ being a latent variable that captures the ability of the candidates (assumed to be a 1D linear scale). For example, $P_c(y=2|x; \alpha)=0.2$ represents that if 1000 candidates at level $\alpha$ attempt the question $x$, then 200 of them would select the second option. 
\vspace{2mm}

\noindent\textbf{Method} We propose to match predicted distributions by using a multiple-choice reading comprehension (MCRC) system $P_{\theta}(y|x)$ trained over standard large-scale MCRC datasets (e.g RACE++). As these systems are trained with cross-entropy loss, their output probabilities should be reflective of model confidence and may correlate with the candidate distributions. One can consider methods to further shape system probabilities, through mass redistribution (Equation \ref{eq:mass_dist}) and temperature annealing (Equation \ref{eq:temperature}).
\begin{align}
    \hat{P}_{\theta}(y|x; \alpha) = (1-\alpha) P_{\theta}(y|x) + \alpha \delta(y, y_{ans}) \label{eq:mass_dist}\\
    \log \hat{P}_{\theta}(y|x; \alpha, \tau) = \frac{\log \hat{P}_{\theta}(y|x; \alpha)}{\tau} \label{eq:temperature}
\end{align}



\noindent We consider using the parameters $\alpha^*, \tau^*$ that ensure that mode accuracy and true class probability of the system and candidates are matched, i.e,
\begin{equation}
    \text{acc}(\hat{P}_{\theta}) = \text{acc}(P_c) \qquad \text{tcp}(\hat{P}_{\theta}) = \text{tcp}(P_c) 
\end{equation}

\noindent This is done at the test level, such that only 2 parameters are found and used to shape all the model probabilities over the entire test set~\footnote{As the parameters $\alpha$ and $\tau$ are optimised for each of the levels of data this may be viewed as an upperbound in performance. However, as only two parameters are being estimated to match all the response distributions for a level, the resulting bias is expected to be small.}. \\

\noindent \textbf{Evaluation} We consider the standard probability distribution divergence metrics as a way to measure the difference between the true candidate and predicted candidate distributions. This includes KL divergence, total variation, and the Hellinger distance. 
\begin{align}
    \text{KL Divergence}   &= \sum_i p_i \log \frac{p_i}{q_i} \\
    \text{Total Variation} &= \frac{1}{2} \sum_i |p_i - q_i| \\
    \text{Hellinger}       &= \frac{1}{\sqrt{2}} \sqrt{\sum_i (\sqrt{p_i} - \sqrt{q_i})^{2}}
\end{align}






\section{Experimental Set-Up}
\noindent \textbf{Training Data} We use RACE++ \cite{pmlr-v101-liang19a} to train and evaluate our base MRC system. RACE++ is a corpora of MCRC exams collected from English exams at middle school, high school and college level for Chinese students. The dataset is one of the largest comprehension corpora available, with the size of splits shown in Table \ref{tab:race++_split}. 
\begin{table}[H]
\small
    \centering
    \begin{tabular}{c|rrr}
        \toprule
        subset & train & valid & test \\
        \midrule
        RACE-M & 25,241  & 1,436 & 1,436 \\
        RACE-H & 62,445  & 3,451 & 3,498 \\
        RACE-C & 12,702  & 712   & 708 \\
        \midrule
        RACE++ & 100,388 & 5,599 & 5,642 \\
        \bottomrule
    \end{tabular}
    \caption{Data splits for RACE++. RACE++ is composed of questions at the middle school (M), high school (H), and college (C) level.}
    \label{tab:race++_split}
\end{table}
\noindent \textbf{Machine System} We take the Electra-Large for the multiple-choice setup described in Section \ref{sec:mcrc} and fine-tune the system on RACE++. We follow the hyperparameters of \citet{raina-gales-2022-answer} and use a batch size of 4, AdamW optimiser, learning rate of $2e^{-6}$, and train for 3 epochs.

\section{Experimental Results}
\begin{figure*}[t]
    \centering
    \begin{subfigure}[t]{0.5\columnwidth}
        \centering
        \includegraphics[width=1.5in]{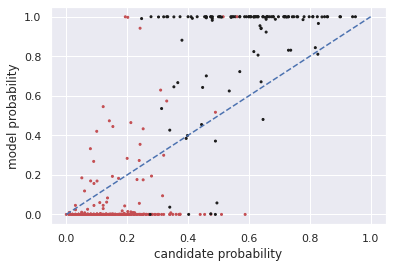}
        \caption{$\tau{=}1, \alpha{=}0$ \hspace{2mm} (\textbf{B1})}
    \end{subfigure}%
    ~ 
    \begin{subfigure}[t]{0.5\columnwidth}
        \centering
        \includegraphics[width=1.5in]{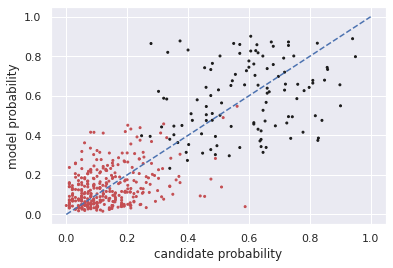}
        \caption{$\tau{=}9.46, \alpha{=}0.044$ \hspace{2mm} (\textbf{B1})}
    \end{subfigure}%
    ~
    \begin{subfigure}[t]{0.5\columnwidth}
        \centering
        \includegraphics[width=1.5in]{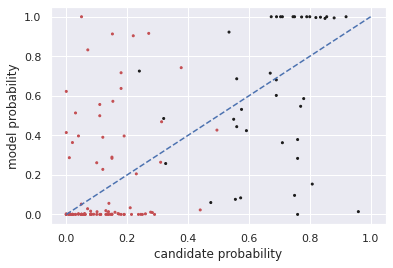}
        \caption{$\tau{=}1, \alpha{=}0$ \hspace{2mm} (\textbf{C2})}
    \end{subfigure}%
    ~ 
    \begin{subfigure}[t]{0.5\columnwidth}
        \centering
        \includegraphics[width=1.5in]{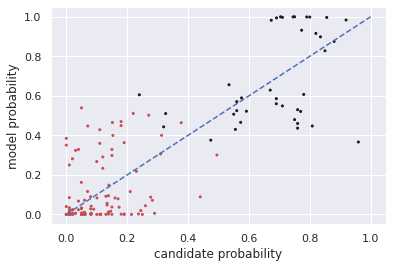}
        \caption{$\tau{=}9.46, \alpha{=}0.044$ \hspace{2mm} (\textbf{C2})}
    \end{subfigure}   
    \caption{Visualisation of model candidate predictions vs true candidate distributions under different shaping parameters for \textbf{B1} and \textbf{C2} tests in CMCQRD.}
    \label{fig:preds_B1_C2}
\end{figure*}

\begin{table}[H]
    \small
    \centering
    \begin{tabular}{c|cc|cc}
        \toprule
                 & \multicolumn{2}{c|}{mode accuracy}  & \multicolumn{2}{c}{true class prob} \\
         test   & cand.  & model                  & cand. & model \\
        \midrule
         RACE++  & -    & 86.8 &  -   & 85.4 \\
         B1      & 91.3 & 90.4 & 59.0 & 88.1  \\ 
         B2      & 90.5 & 73.4 & 60.9 & 70.0 \\
         C1      & 90.3 & 56.9 & 61.3 & 59.1 \\
         C2      & 89.7 & 64.1 & 68.6 & 63.0 \\
        \bottomrule
    \end{tabular}
    \caption{Baseline system performance compared with candidate distributions}
    \label{tab:system_vs_candidates}
\end{table}

\noindent\textbf{Machine Reading Performance} We first consider the portability of a machine reading comprehension system trained on RACE++ to CMCQRD. Table \ref{tab:system_vs_candidates} shows the performance of the MRC system as well as actual candidates, where we display mode accuracies and true class probabilities. The MRC system performs well on RACE++, an ability that transfers over to the learner's exams. However, there is observed performance degradation when the system is used on harder exams, with a 30\% drop in performance from questions at a B1 to C2 level. In contrast, the candidates have similar accuracies across the different tests, as candidates take tests aimed at their ability. Therefore candidates at a higher level will sit the harder exams, while, our MRC system is trained at a singular level. This highlights that for later considerations of candidate distribution matching, one should be able to encode a level of ability, motivating the importance of mass redistribution. 
\vspace{2mm}

\noindent\textbf{Candidate Distribution Matching} We next consider whether MRC systems trained on RACE++ can be leveraged to predict the candidate distributions of CMCQRD. We evaluate both the raw system probabilities and reshaped probabilities, where the reshaping is parameterized by two parameters $\alpha$ and $\tau$ selected at a test level (as described in Section 4). Table \ref{tab:candidate_matching} shows the comparison between the raw system probabilities with the reshaped probabilities, where we observe notable improvements in all the different distance metrics. Visualisations of the predictions versus the true candidate probabilities can be found in Figure \ref{fig:preds_B1_C2}.
\begin{table}[h]
    \small
    \centering
    \begin{tabular}{ccc|cc|ccc}
        \toprule
         & $\tau$  & $\alpha$ & acc & tcp & kl $(\downarrow)$ & h $(\downarrow)$ & v $(\downarrow)$ \\
        \midrule
        \midrule
        \multirow{2}{*}{B1} & 1.0       & 0.00        & 90.4  & 88.1 & 5.04 & 0.87 & 0.37 \\
        & 9.4  & 0.04 & 91.3  & 59.0 & 0.18 & 0.39 & 0.22 \\
        \midrule
        \multirow{2}{*}{B2} & 1.0       & 0.00     & 73.4 & 70.0  & 3.54  & 0.90 & 0.43 \\
        & 5.2 & 0.46 & 90.5 & 60.9  & 0.33  & 0.49 & 0.27 \\
        \midrule
        \multirow{2}{*}{C1} & 1.0       & 0.000      & 56.9  & 59.1 & 3.87  & 1.01 & 0.37 \\
        & 4.5  & 0.49 & 90.3  & 61.3 & 0.44  & 0.57 & 0.30 \\
        \midrule
        \multirow{2}{*}{C2} & 1.0       & 0.000      & 90.4  & 88.1 & 5.04 & 0.87 & 0.37 \\
        & 9.4  & 0.04 & 91.3  & 59.0 & 0.18 & 0.39 & 0.22 \\
        \bottomrule
    \end{tabular}
    \caption{Automatic candidate distribution prediction performance, where the RACE++ trained MRC system is leveraged to predict estimates for CMCQRD.}
    \label{tab:candidate_matching}
\end{table}

\begin{figure}[h]
    \centering
    \includegraphics[width=\columnwidth]{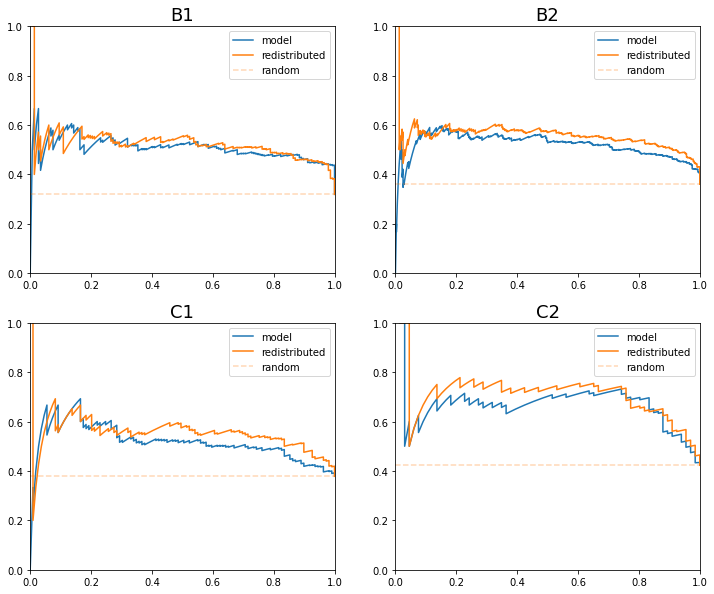}
    \caption{Bad distractor detection precision-recall.}
    \label{fig:distractor_10}
\end{figure}

\vspace{1mm}
\noindent\textbf{Underperforming Distractor Detection}
As a practical use-case of predicting the candidate distributions, we consider the task of bad distractor detection. If a distractor is clearly invalid, such that most or all students realise that the option should not be selected, then this distractor may be one that is of poor quality and could be improved. We define a poor distractor as one that fewer than 10\% of candidates select, and perform a detection task where the lowest probability distractors are selected. Figure \ref{fig:distractor_10} shows the precision-recall (PR) curve for underperforming distractor detection, where we observe that leveraging machine reading comprehension systems trained on external RACE++ data is highly beneficial and leads to performance significantly better than random. 

\section{Conclusions}
In this paper, we analyse CMCQRD, a dataset of multiple choice questions along with actual candidate distributions. We provide general analysis of the properties of the dataset, including ported baseline performance, readability metrics and use of world knowledge. We further consider the task of candidate distribution matching and introduce baseline approaches that can be used to port machine reading comprehension systems trained on RACE++ to generate predicted candidate distributions. Finally, we demonstrate that these systems perform reasonably for real-world tasks such as poor distractor detection. 

\section*{Acknowledgements}
This paper reports on research supported by Cambridge University Press \& Assessment (CUP\&A), a department of The Chancellor, Masters, and Scholars of the University of Cambridge.

\bibliography{anthology,custom}
\bibliographystyle{acl_natbib}

\appendix
\section{Further Analysis}
\subsection{Probability CDFs}
\begin{figure}[h!]
    \centering
    \begin{subfigure}[t]{0.5\columnwidth}
        \centering
        \includegraphics[width=1.4in]{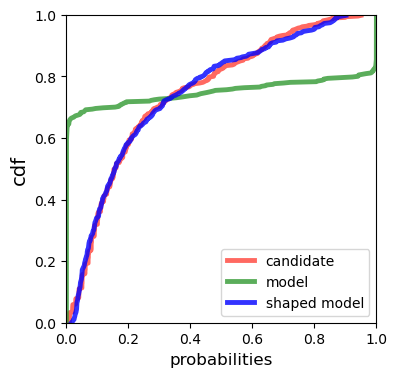}
        \caption{\textbf{B1}}
    \end{subfigure}%
    ~ 
    \begin{subfigure}[t]{0.5\columnwidth}
        \centering
        \includegraphics[width=1.4in]{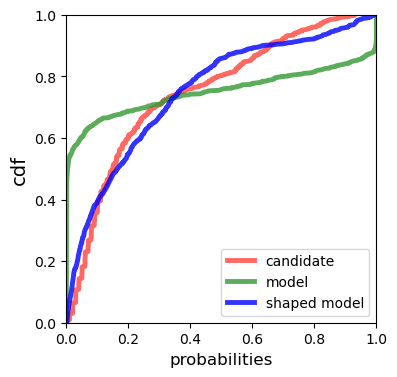}
        \caption{\textbf{B2}}
    \end{subfigure}   
    ~
    \centering
    \begin{subfigure}[t]{0.5\columnwidth}
        \centering
        \includegraphics[width=1.4in]{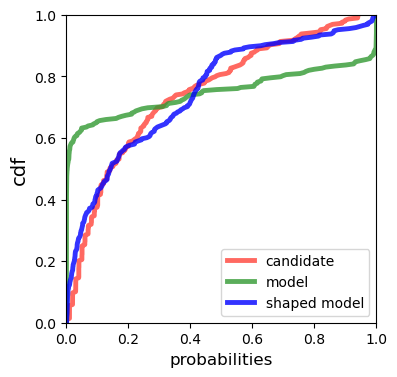}
        \caption{\textbf{C1}}
    \end{subfigure}%
    ~ 
    \begin{subfigure}[t]{0.5\columnwidth}
        \centering
        \includegraphics[width=1.4in]{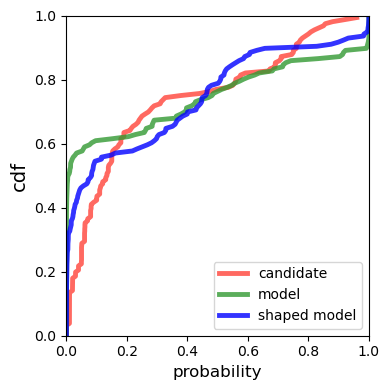}
        \caption{\textbf{C2}}
    \end{subfigure}  
    \caption{Cumulative distribution plot of probabilities of candidate, model and reshaped distributions for all 4 difficulty levels.}
    \label{fig:cdp_of_preds}
\end{figure}

Figure \ref{fig:cdp_of_preds} shows the cumulative distribution function plots over all difficulty levels, where we observe at the B1 level the re-shaped matches the candidate distribution level well, with profiles varying more for the harder exams.  

\subsection{Question Complexity Assessment}
\label{sec:qc_assess}
We analyse the complexity of the dataset using automatic complexity measures. We consider both deep learning classifier based approaches, as well as standard readability metrics: \\

\noindent \textbf{Readability Metrics} $(\uparrow)$: We further consider several other standard readability measures such as the Flesch-Kincaid Grade Level \citep{kincaid1975derivation}, Dale Chall Readability \citep{dale1949concept}, Automated Readability Index (ARI) \citep{senter1967automated}, Coleman Liau Index \citep{coleman1975computer}, Gunning Fog \citep{gunning1952technique}, Spache \citep{spache1953new} and Linsear Write \citep{klare1974assessing}. The greater the readability score, the more challenging the text is to read. \\

\noindent \textbf{Deep classifier} $(\uparrow)$:
A standard transformer-based classifier classifies between \textit{easy}, \textit{medium} and \textit{hard} questions as in \citet{raina2022multiple}. The question, context and answer options are concatenated together into a single input (different forms can be considered by omitting one or more components) which is then passed to the transformer encoder followed by a linear classification head + softmax to return a probability distribution, $[p_{\text{easy}}, p_{\text{medium}}, p_{\text{hard}}]^T$. 
The complexity score is calculated as $ \texttt{complexity} = 0 \times p_{\text{easy}} + 50 \times p_{\text{medium}} + 100 \times p_{\text{hard}} $. The question complexity systems are trained with inputs that are a concatenation of the context, question and all four answer options ($C+Q+\{O\}$). The classifier is a ELECTRA-large transformer with a single linear classification head trained on the training split of RACE++, with a batchsize of 4, learning rate of 2e-6 and for 2 epochs. \\

\begin{table}[h!]
\centering
\begin{small}
    \begin{tabular}{l|cccc}
    \toprule
Score & B1 & B2 & C1 & C2 \\
\midrule
Deep  & $49_{\pm 9}$ & $78_{\pm 22}$ & $96_{\pm 10}$ & $95_{\pm 14}$ \\
\midrule
Flesch-K  & $8.3_{\pm 1.2}$ & $7.9_{\pm 2.0}$ & $11.7_{\pm 1.8}$ & $10.6_{\pm 1.4}$ \\
Dale  & $7.2_{\pm 0.5}$ & $7.5_{\pm 0.7}$ & $9.0_{\pm 0.7}$ & $8.5_{\pm 0.6}$ \\
ARI  & $8.4_{\pm 1.6}$ & $7.9_{\pm 2.6}$ & $12.3_{\pm 2.2}$ & $11.1_{\pm 1.5}$ \\
Coleman  & $8.1_{\pm 1.3}$ & $8.3_{\pm 1.8}$ & $11.7_{\pm 1.5}$ & $10.3_{\pm 1.3}$ \\
Gunning  & $11.3_{\pm 1.2}$ & $10.9_{\pm 2.2}$ & $15.2_{\pm 2.0}$ & $13.8_{\pm 1.7}$ \\
Spache  & $5.5_{\pm 0.5}$ & $5.5_{\pm 0.7}$ & $6.8_{\pm 0.6}$ & $6.5_{\pm 0.5}$ \\
Linsear  & $11.0_{\pm 1.7}$ & $10.0_{\pm 2.8}$ & $14.2_{\pm 2.4}$ & $13.4_{\pm 1.6}$ \\
   \bottomrule
    \end{tabular}
    \end{small}
\caption{Comparing average scores from standard complexity measures across the CMCQRD dataset grade levels.}
\label{tab:read}
\end{table}

\noindent The complexity levels for the CMCQRD dataset is partitioned into B1, B2, C1 and C2 (see Table \ref{tab:camchoice_split}). Table \ref{tab:read} presents the mean difficulty levels on questions from each of these grade levels using several measures of complexity that are based on either a deep classifier or standard readability scores (see Section \ref{sec:qc_assess}). It is observed that all the readability metrics indicate that the readability scores for CEFR levels C1 and C2 are statistically significantly higher than for CEFR levels B1 and B2. However, it is more challenging for these readability metrics to distinguish between B1/B2 or C1/C2. In contrast, the deep learning based classifier is able to further distinguish between the B1 and B2 CEFR levels. The complexity scores for the deep learning classifier are particularly impressive as it is trained explicitly on the RACE++ dataset's interpretation of complexity, demonstrating there is a strong correlation between the grade levels for CMCQRD and RACE++. Given RACE++ gives Middle school level questions a score of 0.0, High school level a score of 0.5 and college level a score of 1.0, the B1 questions can be interpreted to be at High school level, the B2 questions in-between High school and college level, while C1 and C2 questions are both at college level.

\subsection{World Knowledge Assessment} 
Previous work has found that world knowledge can be used as a shortcut when answering MCRC questions, bypassing any required comprehension. \citet{liusie-etal-2023-world} propose to measure world knowledge by training `context-free systems', where an MCRC system is trained with defective inputs $\tilde{x}=(Q, \{O\})$, and by calculating context-free accuracy over a MCRC dataset.

\begin{table}[H]
    \small
    \centering
    \begin{tabular}{c|cc|cc}
        \toprule
                 & \multicolumn{2}{c|}{mode accuracy}  & \multicolumn{2}{c}{true class prob} \\
        test & QOC  & QO   & QOC  & QO \\ 
        \midrule
        RACE++ & 86.8 & 59.4 & 85.4 & 55.2 \\ 
        B1     & 90.4 & 50.4 & 88.1 & 47.5 \\
        B2     & 73.4 & 48.6 & 70.0 & 46.6 \\
        C1     & 56.9 & 51.4 & 59.1 & 45.9 \\
        C2     & 64.1 & 48.7 & 63.0 & 50.1 \\
        \bottomrule
    \end{tabular}
    \caption{Baseline system performance compared with world knowledge systems.}
    \label{tab:world_knowledge}
\end{table}

Table \ref{tab:world_knowledge} demonstrates how much world knowledge can be used to solve the multiple choice questions without context, done by looking at the accuracy of a context-free baseline. We find that without comprehension, one can infer answer information for CMCQRD, however to a less significant degree than for the standard datasets such as RACE++. Additionally, all tests at the different difficulty levels have similar QO accuracies which provides evidence that the increased difficulty in tests does indeed come from harder comprehension as desired.  

\end{document}